\documentclass[journal, letterpaper]{IEEEtran}

\usepackage{graphicx}
\usepackage{url}         
\usepackage{textgreek}	% Greek to me, dawg
\usepackage{listings}
\usepackage{csvsimple}
\usepackage{longtable}

\usepackage[colorlinks=true, allcolors=blue]{hyperref}

\usepackage{natbib}  % DO NOT CHANGE THIS AND DO NOT ADD ANY OPTIONS TO IT
\usepackage{caption} % DO NOT CHANGE THIS AND DO NOT ADD ANY OPTIONS TO IT

\usepackage{algorithm}
\usepackage{algorithmic}
\usepackage{booktabs}
\usepackage{amsmath}
\usepackage{amssymb}
\usepackage{multirow}
\usepackage{authblk}
\usepackage{pifont}
\newcommand{\xmark}{\ding{55}} % 叉号 ✗
\newcommand{\cmark}{\ding{51}} % 对勾 ✓

\title{SpatialAfford: Teaching Compact VLMs Where to Look and Where to Ground for Affordance}

\author[1]{Yufei Zhang}
\author[1]{Chenlu Zhan}
\author[2]{Donghui Sun}
\author[2]{Xiaoxin Chen}
\author[1]{Hongwei Wang*}

\affil[1]{Zhejiang University}
\affil[2]{Vivo Mobile Communication Co., Ltd}

\begin{document}

\maketitle

\begin{abstract}
Affordance grounding aims to localize the functional region for interaction, such as the handle to grasp or the button to press, rather than the whole object. This makes it more challenging than generic visual grounding because the target region is smaller, more ambiguous, and more dependent on task context, especially for compact vision-language models (VLMs) used in embodied settings. Recent sequence-level supervision and reinforcement learning improve coordinate prediction quality, yet compact autoregressive VLMs still lack reliable affordance-aware visual focus before coordinate generation: the model can produce better coordinate tokens while its cross-modal attention remains diffuse and weakly anchored to the true affordance evidence. To address it, we propose SpatialAfford, a two-stage framework that first aligns attention to the ground-truth affordance region through Spatial Attention Alignment (SAA), then refines coordinate prediction with Spatial-Aware GRPO. By explicitly teaching the model where to look before optimizing where to ground, SpatialAfford turns affordance grounding from a purely output-constrained objective into attention-grounded spatial reasoning. Across ShareRobot-Bench, ReasonAff, and PartAfford, SpatialAfford consistently improves affordance grounding, with a compact 4B model outperforming stronger 7B+ baselines.
 \end{abstract}

%==============================================================================
\section{Introduction}
%==============================================================================

Affordance grounding aims to localize the functional region that supports a specified interaction, such as the handle to grasp or the button to press. This capability is central to embodied intelligence because downstream actions depend on identifying \emph{where} interaction should occur, not merely \emph{which} object is present. Compared with generic object grounding, affordance grounding is substantially more demanding: the target region is typically smaller, more visually subtle, and more dependent on task context. A prediction that roughly covers the object may still be unusable for manipulation if it misses the actionable part. Therefore, the core challenge of affordance grounding is to bind the interaction intent to the correct local visual evidence.

Recent vision-language grounding methods have shown that sequence-level supervision and reinforcement learning can improve coordinate generation quality~\citep{vlmr1,visualrft,affordancer1}. Nevertheless, these approaches predominantly constrain the final prediction, while offering little direct supervision on how the model aligns language queries with localized affordance evidence before decoding. This limitation is particularly severe for compact autoregressive VLMs. Owing to limited representational capacity, compressed visual tokens, and diffuse decoder attention, these models are easily distracted by background clutter, object-level shortcuts, or non-functional regions. Recent studies on multimodal reasoning similarly report that visual-token attention can decay during decoding or become dominated by language priors~\citep{lookback,memvr,vise,sayo,directvtoken}. 
% Text-to-image attention in VLMs (e.g., Qwen3-VL-4B; Bai et al., 2025) exhibits only weak-to-moderate alignment with ground-truth affordance regions, even when the model generates plausible coordinates\cite{zhang2026drives}. We refer to this failure as an \textbf{attention--output mismatch}: the model can improve the decoded box without first establishing reliable affordance-aware visual focus. 
Despite predicting accurate bounding boxes, vision-language models like Qwen3-VL-4B \citep{qwen3vl} exhibit an attention sink phenomenon, concentrating attention weights on initial tokens rather than target affordance regions \citep{zhang2026drives}. This causes an attention--output mismatch, where spatial decoding proceeds without anchoring on affordance-relevant visual features.

\begin{figure}[ht]
\centering
\includegraphics[width=\columnwidth]{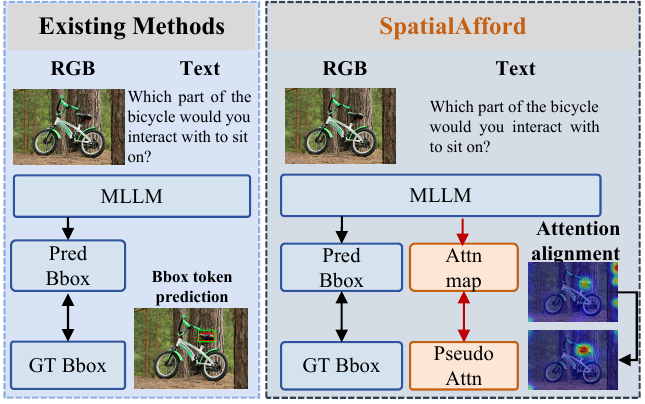}
\caption{Conventional SFT supervises only bbox tokens, whereas SpatialAfford aligns cross-modal attention to the affordance region before bbox prediction.}
\label{fig:teaser}
\end{figure}
To address this bottleneck, we propose \textbf{SpatialAfford}, a two-stage framework that explicitly separates visual evidence alignment from spatial prediction (Figure~\ref{fig:teaser}). Stage~1, \textbf{Spatial Attention Alignment} (SAA), derives a supervision target from the ground-truth affordance box and trains the model to concentrate cross-modal attention within the target region, thereby establishing affordance-aware visual focus. Stage~2, \textbf{Spatial-Aware GRPO}, performs policy optimization on top of these aligned features to refine coordinate prediction. This decomposition is important for compact models: when local evidence is not stabilized first, subsequent coordinate decoding is easily driven by coarse object cues or spurious correlations rather than the true functional region. By aligning the model's visual focus before optimizing its coordinate policy, SpatialAfford turns affordance grounding from a purely output-constrained objective into a visually grounded reasoning process. Our primary contributions are summarized as follows:

\begin{itemize}
    \item \textbf{Problem Diagnosis.} We identify an attention--output mismatch in compact autoregressive VLMs for affordance grounding, showing that coordinate prediction can improve even when cross-modal attention remains misaligned with the local functional region.
    \item \textbf{Two-Stage Framework.} We propose SpatialAfford, which couples Spatial Attention Alignment (SAA) for \emph{where to look} with Spatial-Aware GRPO for \emph{where to ground}, explicitly separating evidence alignment from coordinate refinement.
    \item \textbf{Empirical Results.} We show that this decomposition consistently improves compact affordance grounding across multiple benchmarks, with a 4B model outperforming several larger 7B+ baselines.
\end{itemize}

\section{Related Work}
\label{sec:related_work}

\subsection{Embodied Affordance Grounding and Reinforcement Learning}
\label{sec:rw_affordance_rl}

Affordance grounding~\citep{gibson1979ecological} has transited from foundational convolutional architectures~\citep{umd_affordance,iit_affordance,agd20k,affordancenet,where2act} to decoder-only vision-language models (VLMs)~\citep{llava,llava15,qwen2vl,qwen25vl} that formulate affordance-region grounding as autoregressive coordinate generation. Building upon open-vocabulary detection and embodied foundation models~\citep{mdetr,glip,owlvit,groundingdino,rt2_arxiv,palm_e_arxiv,saycan_arxiv,openvla_arxiv}, general-purpose embodied systems such as RoboBrain~\citep{robobrain,robobrain2,robobrain25} and RynnBrain~\citep{rynnbrain} adopt a two-stage paradigm that pairs supervised instruction tuning with subsequent policy optimization for broad spatial manipulation. In parallel, reinforcement learning including RLHF~\citep{rlhf,llava_rlhf,rlhf_v} and GRPO~\citep{grpo} have been extended to spatial grounding tasks~\citep{visualrft,vlmr1,r1v,mgpo,dapo}. Specifically, Affordance-R1~\citep{affordancer1} leverages GRPO to optimize coordinate token generation for affordance reasoning.

However, existing embodied VLM training and policy optimization paradigms compute supervisory loss gradients exclusively on output coordinate tokens. Consequently, compact models tend to optimize coordinate prediction through global object shortcuts while their internal cross-modal attention remains unaligned with actionable affordance regions. SpatialAfford decouples this optimization by explicitly aligning internal cross-modal focus with affordance regions during Stage 1 SFT, establishing a rectified feature space before performing Stage 2 policy optimization.

\subsection{Attention Regularization and Supervision}
\label{sec:rw_attention}

Cross-modal alignment in autoregressive VLMs depends critically on visual token attention. Prior diagnostic studies confirm that directing visual attention enhances recognition efficiency~\citep{lookwhere}, whereas supervising answer-to-visual-token attention improves grounding accuracy~\citep{directvtoken}. Related studies further emphasize the necessity of visual retracing during decoding~\citep{memvr,lookback} and demonstrate that explicit attention regularization stabilizes multimodal reasoning~\citep{sayo,vise}. Furthermore, TokAG~\citep{tokag} reveals that internal cross-modal attention intrinsically harbors functional affordance cues through training-free heatmap extraction. SpatialAfford converts intermediate cross-modal attention into an explicit, region-level supervisory target during training, forcing compact models to anchor visual focus onto actionable sub-regions prior to coordinate emission.

\begin{figure*}[ht]
\centering
\includegraphics[width=\textwidth]{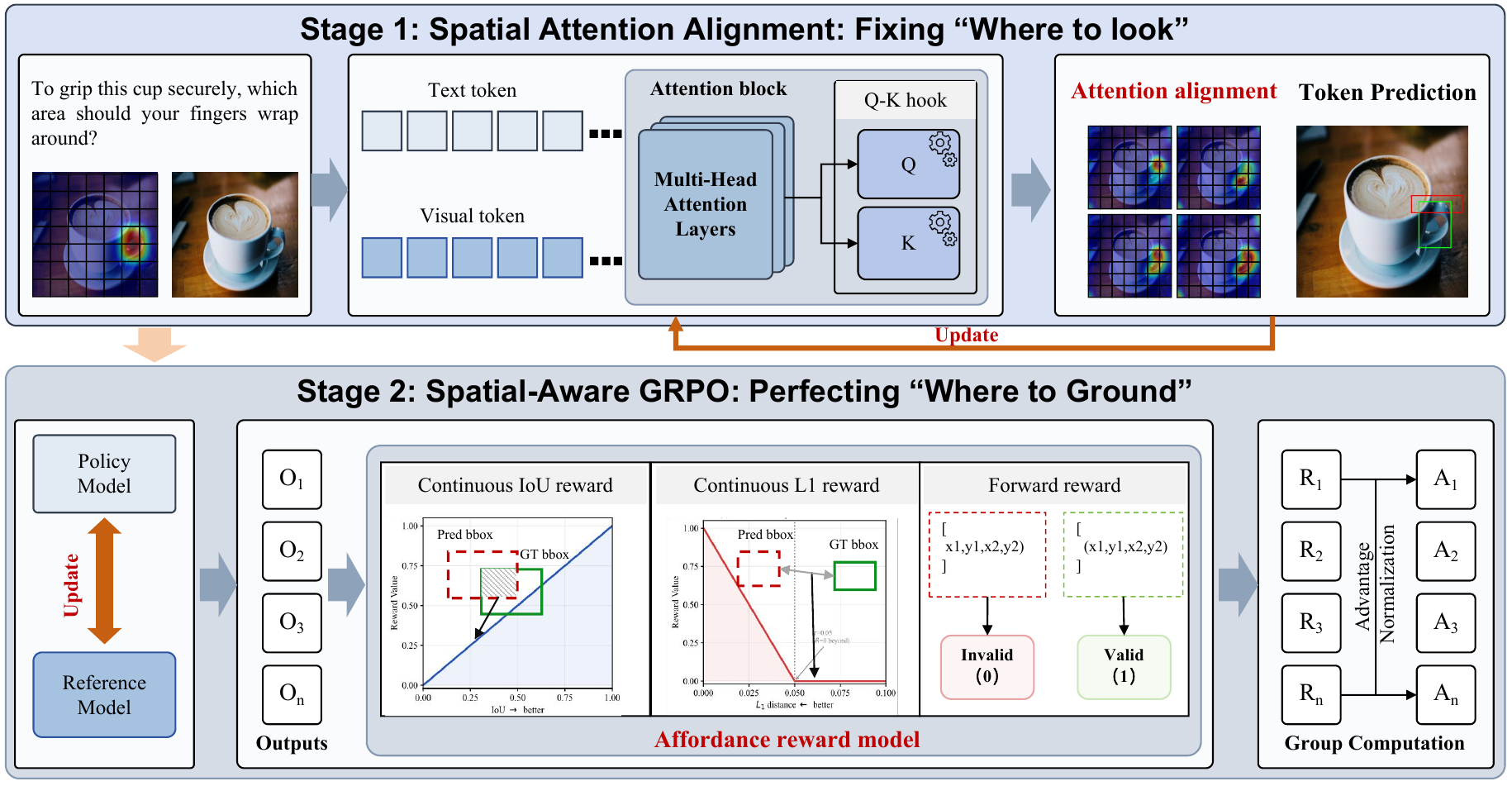}
\caption{Overall framework of SpatialAfford. \textbf{Stage~1:} Spatial Attention Alignment aligns the model's visual focus with the affordance region, establishing spatial-semantic correspondence between the instruction and image and resolving \emph{where to look}. \textbf{Stage~2:} Spatial-Aware GRPO improves bounding box prediction with spatial rewards, resolving \emph{where to ground} for precise affordance grounding.}
\label{fig:framework}
\end{figure*}
%==============================================================================
\section{Methodology}
\label{sec:method}
%==============================================================================
The overall framework is illustrated in Figure~\ref{fig:framework}. To address the attention--output mismatch, SpatialAfford adopts a coarse-to-fine training strategy. Stage~1, \emph{Spatial Attention Alignment (SAA)}, explicitly aligns text-to-image attention with the ground-truth affordance region, so the model first learns \emph{where to look}. Stage~2, \emph{Spatial-Aware GRPO}, then optimizes coordinate generation on top of these aligned features, so the model learns \emph{where to ground}. This decomposition turns affordance grounding into a sequential process of affordance-evidence alignment followed by spatial coordinate refinement.
\subsection{Overview and Conceptual Formulation}
\label{sec:overview}
We formalize the \emph{attention--output mismatch} in decoder-only VLMs. Given an input image $\mathbf{I} \in \mathbb{R}^{H \times W \times 3}$ and a task instruction $T$ (e.g., ``reach the mug's handle''), a VLM autoregressively predicts a bounding box $\mathbf{b} = (x_{\min}, y_{\min}, x_{\max}, y_{\max})$ for the targeted affordance region. The network processes the input via a vision encoder that generates $N_{\text{img}}$ visual patch tokens, coupled with a language backbone of $L$ layers wherein the $l$-th layer computes a cross-modal attention matrix $\mathbf{A}^{(l)} \in \mathbb{R}^{N_{\text{text}} \times N_{\text{img}}}$ mapping text query tokens to visual patch tokens.

\textbf{Attention--output mismatch.} Standard supervised fine-tuning (SFT) optimizes solely the negative log-likelihood over the sequence of coordinate tokens:
\begin{equation}
\mathcal{L}_{\text{SFT}} = -\log P(\mathbf{b}_{\text{gt}} \mid \mathbf{I}, T) = -\sum_{t} \log p(b_t \mid b_{<t}, \mathbf{I}, T).
\label{eq:sft}
\end{equation}
Although gradients backpropagate through intermediate layers during SFT, this cross-entropy objective provides no explicit supervisory signal on the cross-modal attention matrix $\mathbf{A}^{(l)}$. Lacking direct spatial regularization, implicit gradients are insufficient to anchor visual focus, causing attention mass to diffuse across task-irrelevant regions. 

Let $\mathcal{M}_{\text{gt}} \subset \{1, \dots, N_{\text{img}}\}$ denote the set of visual token indices corresponding to the spatial interior of the ground-truth region $\mathbf{b}_{\text{gt}}$. Instantiating this framework with Qwen3-VL-4B as our primary backbone, diagnostic empirical evaluations reveal that the normalized attention density within $\mathcal{M}_{\text{gt}}$ remains suppressed, accompanied by near-zero or negative Normalized Scanpath Saliency (NSS) scores (Appendix~\ref{sec:appendix_attn}). We formalize this empirical imbalance as:
\begin{equation}
\mathbb{E}_{i \in \mathcal{M}_{\text{gt}}}\!\left[\mathbf{A}_{j,i}\right]
<
\mathbb{E}_{i \notin \mathcal{M}_{\text{gt}}}\!\left[\mathbf{A}_{j,i}\right],
\label{eq:broken}
\end{equation}
where $\mathbf{A}_{j,i}$ represents the aggregated cross-modal attention weight from query token $j$ to visual patch token $i$. This observed \emph{attention--output mismatch} suggests that coordinate generation is susceptible to shortcut learning, failing to ground its internal visual focus on affordance evidence.

\subsection{Spatial Attention Alignment: Fixing ``Where to Look''}
\label{sec:acm}

Given a ground-truth affordance bounding box $\mathbf{B} = (x_1, y_1, x_2, y_2)$ in normalized coordinates, Spatial Attention Alignment (SAA) constructs a spatial target distribution $P_{\mathbf{M}}$ over the visual token grid to provide explicit cross-modal attention supervision.

\textit{Step 1: Bbox to grid mask.} The bounding box $\mathbf{B}$ is mapped onto the visual feature grid of dimension $H_g \times W_g$, dynamically determined by the spatial resolution of the vision encoder:
\begin{equation}
\mathbf{M}[i, j] = \begin{cases} 1 & \text{if } (i, j) \in \text{grid}(\mathbf{B}) \\ 0 & \text{otherwise} \end{cases}
\end{equation}
where $\text{grid}(\mathbf{B})$ maps norm1k coordinates to grid indices via $g_x = \lfloor x/1000 \cdot W_g \rfloor$.

\textit{Step 2: Box blur smoothing.} We smooth the binary mask with a uniform average filter whose kernel size is adaptive to the bbox dimensions:
\begin{equation}
k = \max\!\left(3,\; \min\!\left(\left\lfloor \frac{\max(b_w, b_h)}{2} \right\rfloor \times 2 + 1,\; 7\right)\right)
\end{equation}
where $b_w, b_h$ are the bbox width and height in grid cells. The smoothing window scales with the target extent and is constrained to the odd-sized range of $3$ to $7$ in order to preserve locality on the visual token grid. The kernel $\frac{1}{k^2}\mathbf{1}_{k \times k}$ is applied via 2D convolution:
\begin{equation}
\tilde{\mathbf{M}} = \mathrm{Conv2D}\!\left(\mathbf{M},\; \tfrac{1}{k^2}\mathbf{1}_{k \times k},\; \text{padding}{=}k/2\right)
\end{equation}

\textit{Step 3: Normalization.} The smoothed mask is normalized into a probability distribution:
\begin{equation}
P_{\mathbf{M}} = \frac{\tilde{\mathbf{M}}}{\sum_{i,j} \tilde{\mathbf{M}}[i,j]}
\end{equation}
The resulting $P_{\mathbf{M}}$ is used as the supervision target and is derived directly from the geometry of the affordance annotation.

To enforce spatial focus, we minimize the KL divergence $\mathrm{KL}(P_{\mathbf{S}} \| P_{\mathbf{M}})$, penalizing probability mass assigned by the model's attention distribution $P_{\mathbf{S}}$ to task-irrelevant regions where $P_{\mathbf{M}}$ approaches zero. Formally, let $P_{\mathbf{S}}$ denote the normalized text-to-image cross-modal attention map, aggregated across text instruction tokens and designated transformer layers $\mathcal{L}$. The SAA loss is formulated as:

\begin{equation}
\mathcal{L}_{\text{SAA}} = \mathrm{KL}\!\left(P_{\mathbf{S}} \;\|\; P_{\mathbf{M}}\right) = \sum_{(u,v)} P_{\mathbf{S}}[u,v] \log\frac{P_{\mathbf{S}}[u,v]}{P_{\mathbf{M}}[u,v] + \epsilon}
\label{eq:acm_loss}
\end{equation}

where $\epsilon > 0$ is a small constant ensuring numerical stability. Cross-modal attention is extracted exclusively over instruction prompt positions to prevent teacher-forcing shortcut leakage during training. The overall objective balances standard language modeling with attention alignment.
\begin{equation}
\mathcal{L}_{\text{Stage1}} = \mathcal{L}_{\text{SFT}} + \alpha \cdot \mathcal{L}_{\text{SAA}}
\label{eq:stage1}
\end{equation}

\subsection{Spatial-Aware GRPO: Perfecting ``Where to Ground''}
\label{sec:grpo}
While supervised attention alignment establishes affordance-aware features $\mathcal{F}_{\text{SAA}}$, policy optimization sharpens affordance grounding by directly refining coordinate emission over this rectified feature representation.

\subsubsection{Group Relative Policy Optimization}
We adopt Group Relative Policy Optimization~\citep{grpo} for reward-based boundary refinement. For each query, the policy $\pi_\theta$ samples $G$ candidate outputs $\{o_1, \ldots, o_G\}$, and the sampled responses are evaluated by a spatial reward function. The group-relative advantage is computed as:
\begin{equation}
\hat{A}_g = \frac{R_g - \bar{R}}{\sigma_R + \epsilon}, \quad \bar{R} = \frac{1}{G}\sum_{j=1}^{G} R_j
\label{eq:grpo_adv}
\end{equation}
where $R_g$ is the reward of the $g$-th sampled output, and $\sigma_R$ denotes the standard deviation of group rewards.

The policy parameters are updated via a clipped surrogate objective regularized by a reference policy $\pi_{\text{ref}}$:
\begin{equation}
\begin{aligned}
\mathcal{L}_{\text{GRPO}} =
\frac{1}{G}\sum_{g=1}^{G}
\Big[
&\min\!\Big(
\rho_g \hat{A}_g,\;
\mathrm{clip}(\rho_g, 1-\epsilon, 1+\epsilon)\hat{A}_g
\Big) \\
&- \beta D_{\mathrm{KL}}(\pi_\theta \| \pi_{\text{ref}})
\Big],
\end{aligned}
\label{eq:grpo_obj}
\end{equation}
where $\rho_g = \frac{\pi_\theta(o_g \mid \mathbf{I}, T)}{\pi_{\text{ref}}(o_g \mid \mathbf{I}, T)}$ is the policy ratio. The clipped term stabilizes policy updates, while the KL penalty keeps the optimized policy close to the initialization. We use the BNPO loss variant~\citep{dapo} with adaptive KL control ($\beta \in [0.10, 0.50]$, $\text{KL}_{\text{target}}{=}1.0$) for training stability.

\subsubsection{Spatial Reward Engine}
To improve affordance grounding, we design a composite spatial reward function comprising three complementary components: spatial overlap reward ($R_{\text{IoU}}$), coordinate proximity reward ($R_{L_1}$), and format reward ($R_{\text{format}}$).

\textbf{Spatial overlap reward.} To evaluate affordance-region overlap, we compute the Intersection over Union (IoU) between the predicted bounding box $\mathbf{b}(o_g)$ parsed from response $o_g$ and the ground-truth annotation $\mathbf{b}_{\text{gt}}$:
\begin{equation}
R_{\text{IoU}} = \frac{1}{\max(M, N)} \sum_{(i,j) \in \mathcal{M}^*} \mathrm{IoU}(\mathbf{b}_i, \mathbf{b}_j^*)
\end{equation}
where $\mathcal{M}^*$ is the optimal assignment between $M$ predicted and $N$ ground-truth boxes.

\textbf{Coordinate proximity reward.} To further guide fine-grained coordinate adjustment, we compute the mean absolute distance between predicted and ground-truth box coordinates in the norm1k coordinate space:
\begin{equation}
\begin{aligned}
R_{L_1} &=
\max\!\left(0,\; 1 -
\frac{\bar{d}_1(\mathbf{b}(o_g), \mathbf{b}_{\text{gt}})}{\tau_{L_1}}\right),\\
\tau_{L_1} &= 0.05 \cdot \frac{W + H}{2}.
\end{aligned}
\end{equation}
where $\mathbf{b}(o_g)$ denotes the bbox parsed from sampled output $o_g$, $\bar{d}_1 = \frac{1}{4}\sum_{k=1}^{4}|b_k(o_g) - b_{\text{gt},k}|$ is the mean absolute coordinate error, and $W=H=1000$ for norm1k coordinates.

\textbf{Format reward.} To ensure that the model's outputs strictly adhere to the required coordinate format $[(x_1, y_1, x_2, y_2)]$, a format reward is incorporated into the training framework. Specifically, a binary indicator $R_{\text{format}} \in \{0, 1\}$ is defined to validate syntax and parsing integrity:
\begin{equation}
R_{\text{format}} = \begin{cases} 1, & \text{if } \mathbf{b}(o_g) \text{ is valid,} \\ 0, & \text{otherwise.} \end{cases}
\end{equation}

\textbf{Combined reward:} The overall reward function combines spatial accuracy with structural constraints:
\begin{equation}
R = \frac{R_{\text{IoU}} + R_{L_1}}{2} + R_{\text{format}}
\label{eq:reward}
\end{equation}
where $R_{\text{IoU}}, R_{L_1} \in [0, 1]$ are the matched IoU and $L_1$ rewards, and $R_{\text{format}} \in \{0,1\}$ checks valid bbox syntax. The two spatial terms are equally weighted. Since our training data contains single-bbox annotations, the spatial reward $(R_{\text{IoU}} + R_{L_1})/2$ provides the primary learning signal.

\begin{figure*}[ht]
    \centering
    \includegraphics[width=\linewidth]{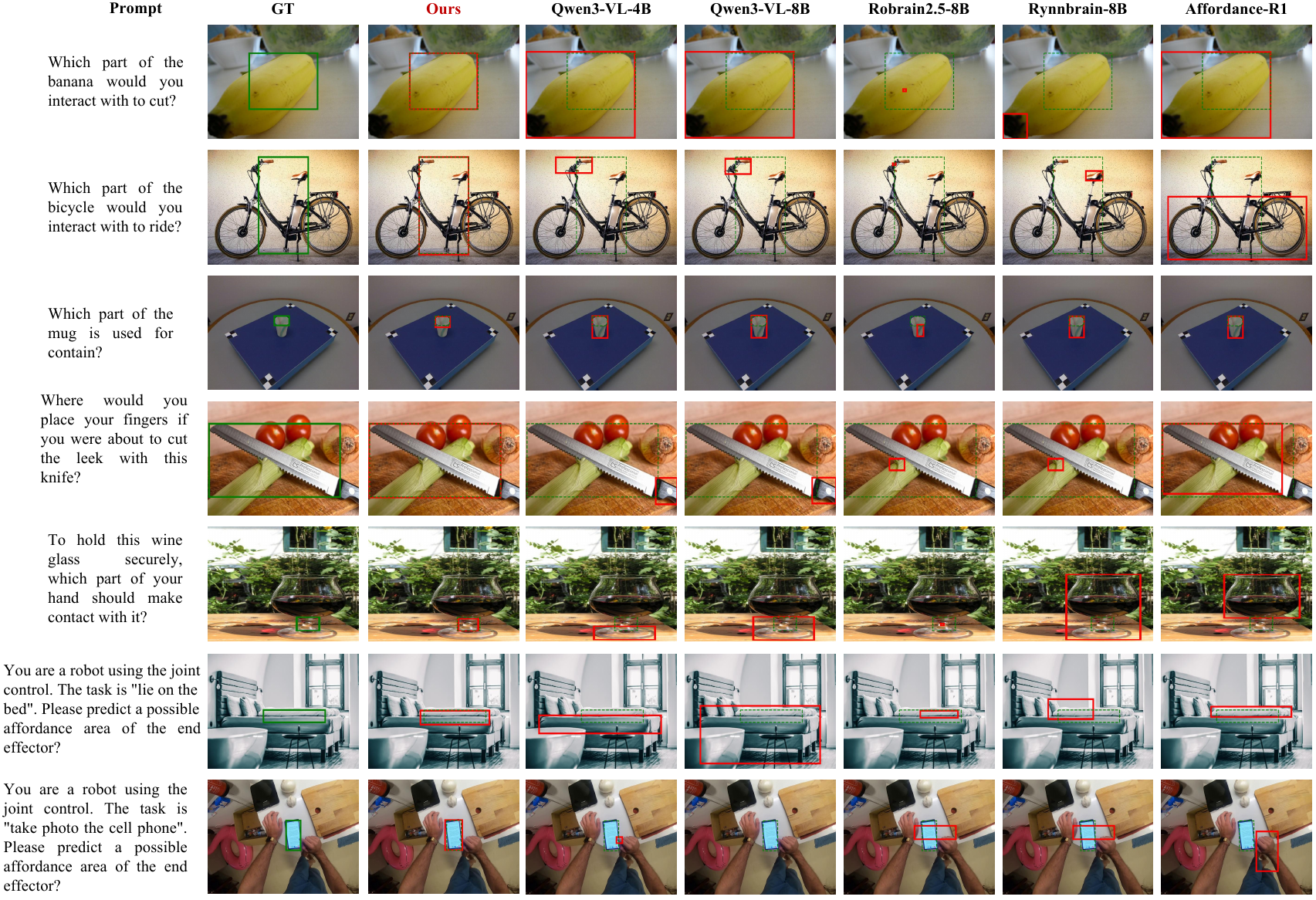}
    \caption{Qualitative comparison of bounding box predictions. Green: ground truth; Red: predict result. The baseline model exhibits severe affordance grounding errors, while SpatialAfford produces accurate predictions.}
    \label{fig:qualitative}
\end{figure*}

\subsection{Synergistic Training: Asymmetric Two-Stage Dependency}
\label{sec:synergy}
Stage~1 performs full-parameter SFT with $\mathcal{L}_{\text{Stage1}} = \mathcal{L}_{\text{SFT}} + \alpha \cdot \mathcal{L}_{\text{SAA}}$ to produce $\mathcal{F}_{\text{SAA}}$, followed by Stage~2 full-parameter GRPO on $\mathcal{F}_{\text{SAA}}$ for coordinate refinement.

\textbf{Stage~1 as the prerequisite.} The policy optimization in Stage~2 is conditioned on the rectified feature space $\mathcal{F}_{\text{SAA}}$:
\begin{equation}
\max_\theta \; \mathbb{E}_{\mathbf{b} \sim \pi_\theta(\cdot \mid \mathcal{F}_{\text{SAA}}(\mathbf{I}, T))} \left[ R(\mathbf{b}, \mathbf{b}_{\text{gt}}) \right]
\label{eq:stage2_conditional}
\end{equation}
Without $\mathcal{F}_{\text{SAA}}$, the reward is applied to scattered or task-irrelevant visual evidence, making policy optimization less effective for precise affordance grounding.

\textbf{Stage~2 as the refinement.} Spatial Attention Alignment alone does not determine precise box boundaries. GRPO further optimizes the mapping from aligned visual features to coordinate outputs, completing the transition from \emph{where to look} to \emph{where to ground}.

%==============================================================================
\section{Experiments}
%==============================================================================

\subsection{Experimental Settings}

\textbf{Training Data.} We compile a unified affordance grounding dataset from three primary sources: \textbf{ShareRobot-Affordance}~\citep{robobrain}, \textbf{ReasonAff-Bbox}~\citep{affordancer1}, and \textbf{PartAfford-Tools}~\citep{partafford}. This combination encompasses robotic manipulation scenarios, complex reasoning-driven affordance descriptions, and part-level functional annotations. Further details are elaborated in Appendix~\ref{sec:appendix_data}.

\textbf{Out-of-Distribution Benchmark.} To evaluate zero-shot generalizability, we benchmark on \textbf{AGD20K}~\citep{agd20k}. AGD20K serves as a rigorous out-of-distribution (OOD) testbed due to its distinct egocentric perspectives and non-overlapping category splits relative to the training corpora. Moreover, its pixel-level affordance annotations enable cross-modal attention alignment evaluation.

\textbf{Baselines.} We compare with three groups of methods: (1)~general-model VLMs, including Qwen2.5-VL-3B/7B~\citep{qwen25vl}, Qwen3-VL-4B/8B~\citep{qwen3vl}, and UniVG-R1~\citep{univg}; (2)~embodied VLMs, including RoboBrain2.0-3B/7B~\citep{robobrain2}, RoboBrain2.5-4B/8B~\citep{robobrain25}, and RynnBrain-2B/8B~\citep{rynnbrain}; and (3)~affordance-specific VLMs AffordanceLLM~\cite{qian2024affordancellm}, Affordance-R1~\citep{affordancer1}.

\textbf{Metrics and Implementation.} We evaluate grounding performance using standard benchmarks, including generalized IoU (gIoU), Precision at IoU threshold $0.5$ ($\text{P@50}$), and syntactic format compliance rate. SpatialAfford is instantiated on Qwen3-VL-4B~\citep{qwen3vl} via a two-stage optimization pipeline: full-parameter supervised fine-tuning with Spatial Attention Alignment (SAA) loss, followed by full-parameter Group Relative Policy Optimization (GRPO)~\citep{grpo} with a group size of $G=4$. Comprehensive training configurations and hyperparameter specifications are detailed in Appendix~\ref{sec:appendix_hparams}.

\subsection{Main Results}
\label{sec:exp_main}

The experimental evaluations validate the effectiveness and generalizability of SpatialAfford. We benchmark across ShareRobot-Bench, ReasonAff, and PartAfford to systematically assess affordance grounding performance.

\begin{table}[ht]
\centering
\caption{Affordance grounding results on ShareRobot-Bench and PartAfford. All metrics are percentages.}
\label{tab:sharerobot_compare}
\resizebox{\linewidth}{!}{
\begin{tabular}{l c c c c c c c}
\toprule
\multirow{2}{*}{Model} & \multirow{2}{*}{Size} & \multicolumn{3}{c}{ShareRobot-Bench} & \multicolumn{3}{c}{PartAfford} \\
\cmidrule(lr){3-5} \cmidrule(lr){6-8}
 & & gIoU$\uparrow$ & cIoU $\uparrow$ & P@50$\uparrow$ & gIoU$\uparrow$ & cIoU$\uparrow$ & P@50$\uparrow$ \\
\midrule
General-purpose VLMs & & & & & & \\
Qwen2.5-VL & 3B & 18.60 & 15.08& 7.00 & 41.29 & 32.31 & 35.31 \\
Qwen3-VL & 4B & 25.10 & 17.59& 17.68 & 47.37 & 45.00 & 46.21 \\
Qwen2.5-VL & 7B & 22.83 & 22.60 & 13.19 & 47.20 & 40.30 & 43.36 \\
Qwen3-VL & 8B & 37.00 & 24.34 & 18.00 & 56.55 & \underline{55.51} & 59.48 \\
\midrule
Embodied VLMs & & & & & & & \\
RoboBrain2.0 & 3B & 28.01 & 15.85 & 26.02 & 38.05 & 36.72 & 35.78 \\
RoboBrain2.5 & 4B & 41.78 & 35.19 & 41.44 & 50.39 & 51.27 & 54.98 \\
RynnBrain & 4B & 43.30 & 42.51 & 37.50 & 35.71 & 33.32 & 35.31 \\
RoboBrain2.0 & 7B & 46.60 &33.67 & 48.50 & 38.86 & 37.09 & 28.44 \\
Affordance-R1 & 7B & 29.44 & 28.30 & 17.00 & \underline{57.24} & 54.95 & \underline{60.90} \\
RoboBrain2.5 & 8B & \textbf{48.07} & \textbf{54.35} & \underline{51.80} & 58.38 & 58.34 & 60.12 \\
RynnBrain & 8B & 44.70 & 40.05 & 40.30 & 38.47 & 35.66 & 35.78 \\
\midrule
\textbf{SpatialAfford} & \textbf{4B} & \underline{47.80} & \underline{49.76} & \textbf{55.00} & \textbf{76.37} & \textbf{78.50} & \textbf{86.57} \\
\bottomrule
\end{tabular}
}
\end{table}

\begin{table}[ht]
\centering
\caption{Affordance reasoning on ReasonAff.}
\label{tab:reasonaff_compare}
\resizebox{\linewidth}{!}{
\begin{tabular}{l c c c c c}
\toprule
Model & Size & gIoU$\uparrow$ & cIoU$\uparrow$ & P@50$\uparrow$ & P@50--95$\uparrow$ \\
\midrule
Grounding methods & & & & & \\
VLPart & --- & 4.21 & 3.88 & 1.31 & 0.85 \\
OVSeg & --- & 16.52 & 10.59 & 9.89 & 4.12 \\
SAN & --- & 10.21 & 13.45 & 7.18 & 3.17 \\
LISA & 7B & 38.17 & 40.58 & 33.62 & 19.69 \\
SAM4MLLM & 7B & 45.51 & 33.64 & 43.48 & 22.79 \\
Seg-Zero & 7B & 59.26 & 48.03 & 61.33 & 45.87 \\
Vision Reasoner & 7B & 63.04 & 52.70 & 67.33 & 47.23 \\
\midrule
General-purpose VLMs & & & & & \\
Qwen3-VL & 4B & 42.88 & 36.81 & 41.17 & 28.50 \\
Qwen2.5-VL & 7B & 25.18 & 20.54 & 26.00 & 15.82 \\
InternVL3 & 8B & 31.79 & 24.68 & 35.41 & 21.93 \\
Qwen3-VL & 8B & 48.40 & 41.20 & 49.83 & 35.20 \\
\midrule
Embodied VLMs & & & & & \\
RynnBrain & 2B & 42.80 & 39.88 & 42.50 & 28.73 \\
RoboBrain2.0 & 3B & 26.26 & 26.12 & 22.33 & 0.11 \\
RoboBrain2.5 & 4B & 26.76 & 35.60 & 24.79 & 9.67 \\
RoboBrain2.0 & 7B & 47.69 & 48.98 & 46.83 & 36.86 \\
RynnBrain & 8B & 49.70 & 43.13 & 49.83 & 37.53 \\
RoboBrain2.5 & 8B & 41.38 & 35.97 & 42.50 & 22.83 \\
AffordanceLLM & 7B & 48.49 & 38.61 & 42.11 & 20.19 \\
Affordance-R1 & 7B & \underline{67.41} & \underline{62.72} & \underline{74.50} & \underline{55.22} \\
\midrule
\textbf{SpatialAfford} & \textbf{4B} & \textbf{74.92} & \textbf{69.32} & \textbf{82.50} & \textbf{64.42} \\
\bottomrule
\end{tabular}
}
\end{table}

Table~\ref{tab:sharerobot_compare} presents the affordance grounding results on ShareRobot-Bench and PartAfford. SpatialAfford achieves superior performance across both benchmarks. On ShareRobot-Bench, SpatialAfford (4B) achieves 55.00\% P@50 and 47.80\% gIoU, surpassing RoboBrain2.5-8B (51.80\% P@50) while nearly matching its gIoU (48.07\%). Table~\ref{tab:reasonaff_compare} summarizes the performance on ReasonAff, which evaluates instruction-dependent affordance reasoning. SpatialAfford achieves 74.92\% gIoU and 82.50\% P@50, outperforming the affordance-specialized Affordance-R1 by +7.51\% gIoU and +8.00\% P@50. These improvements demonstrate that enforcing spatial focus prior to policy optimization remains effective when functional targets must be inferred from contextual language instructions.

\begin{figure}[ht]
\centering
\includegraphics[width=\linewidth]{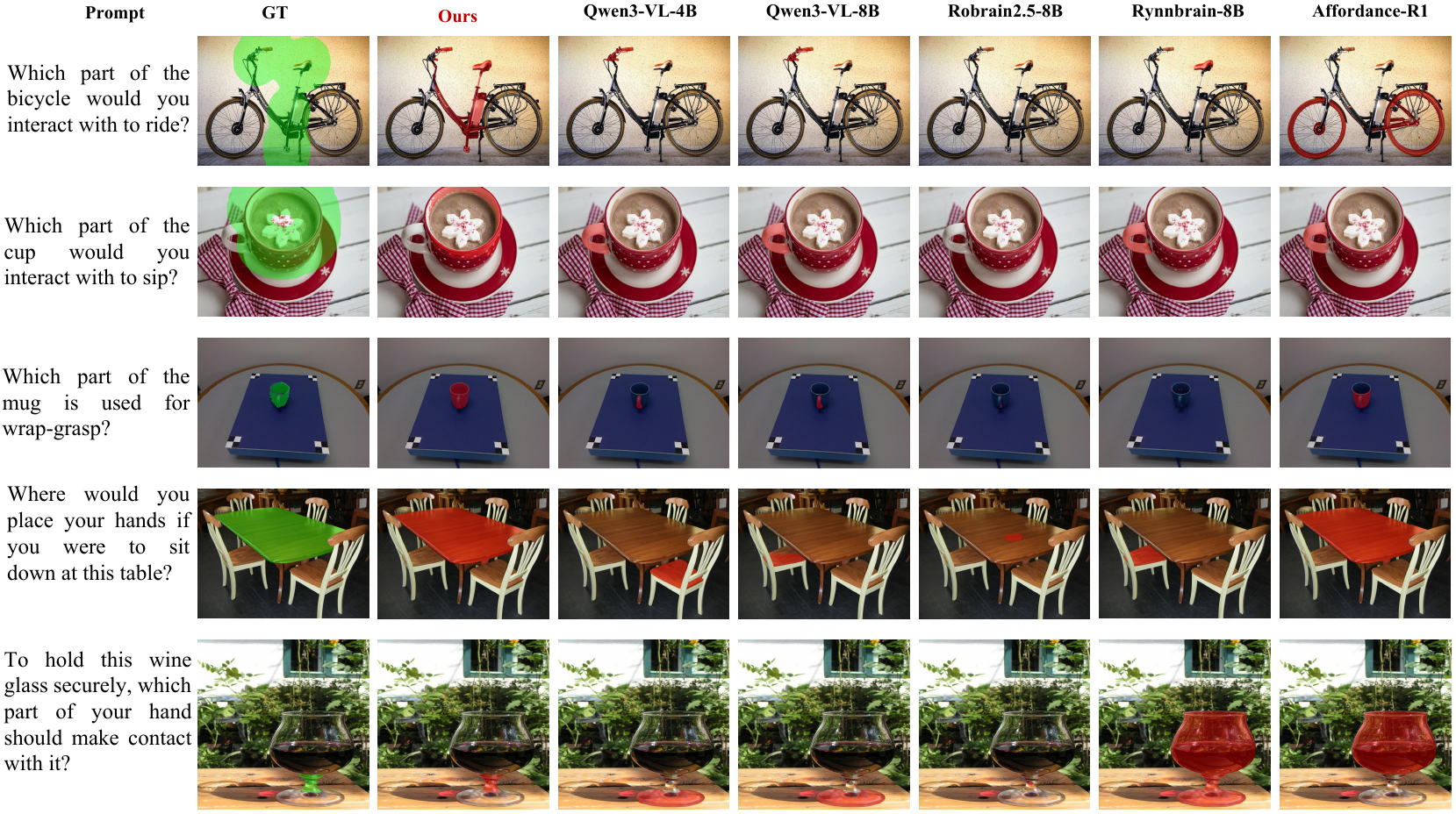}
\caption{Qualitative comparison of affordance mask predictions. SpatialAfford generates precise masks that align with ground truth affordance regions.}
\label{fig:mask}
\end{figure}

Figures~\ref{fig:qualitative} and~\ref{fig:mask} provide qualitative visualizations of predicted affordance-region boxes and cross-modal attention distributions. Competing methods frequently suffer from object-level shortcuts or diffuse over adjacent non-functional structures. In contrast, SpatialAfford consistently anchors visual focus onto actionable affordance regions, yielding precise affordance grounding that aligns with the quantitative metrics.

\subsection{Generalization and Attention Analysis}
\label{sec:exp_attn}
To evaluate the generalization ability of SpatialAfford, we present zero-shot affordance reasoning results on AGD20K in Table~\ref{tab:attn}. At a compact 4B scale, SpatialAfford achieves top scores in P@50 (37.44\%)and NSS (1.05), while ranking second only to the 8B RoboBrain2.5 in gIoU and cIoU. These diagnostics confirm its cross-scene generalization and justify the necessity of SAA to guide models on \emph{where to look} prior to coordinate regression.

\begin{figure*}[ht]
\centering
\includegraphics[width=1\linewidth]{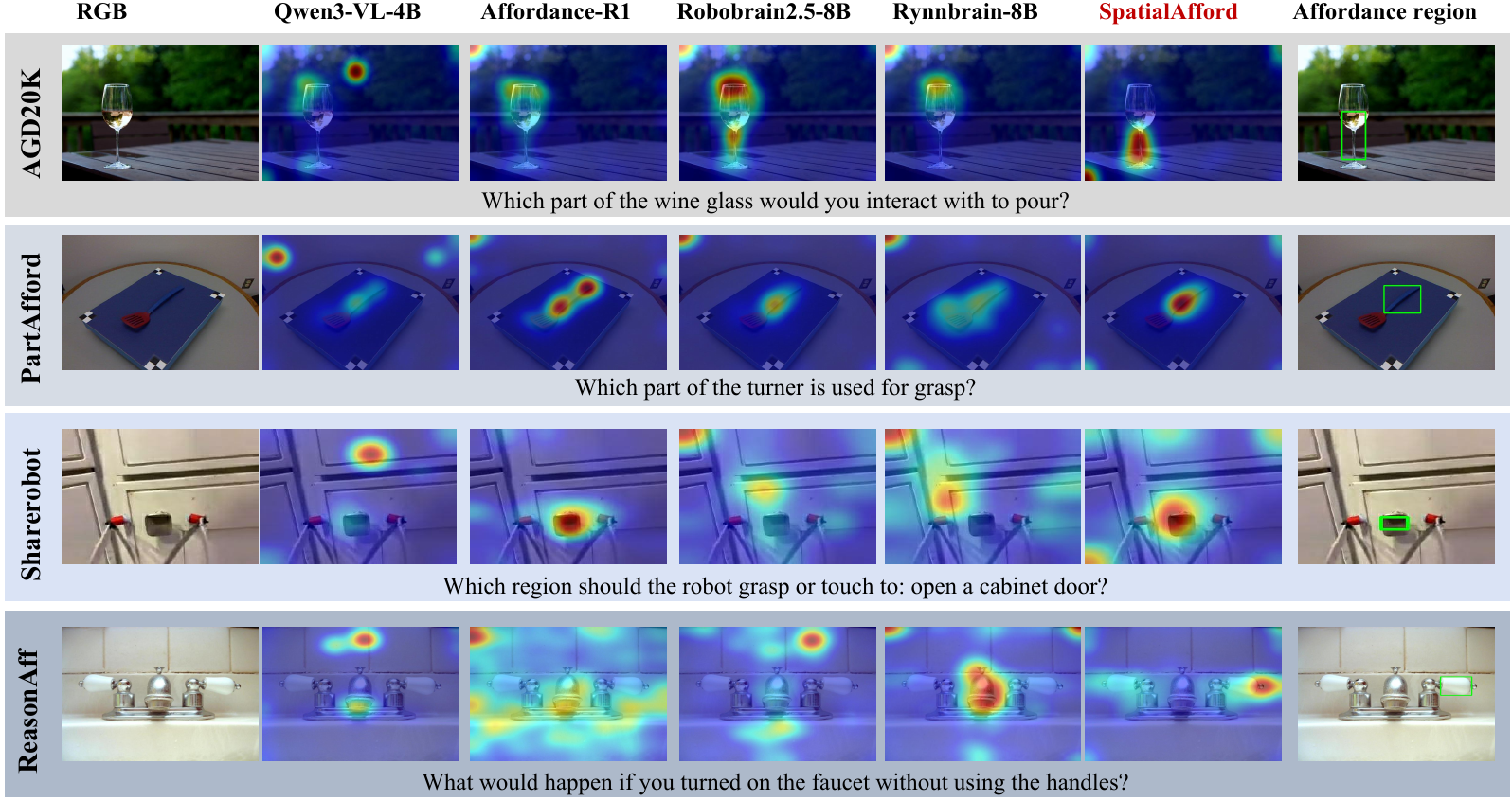}
\caption{Cross-modal attention maps for affordance grounding. The visualization covers different evaluated datasets and supports inspection of whether model attention concentrates on affordance regions or diffuses over irrelevant background areas.}
\label{fig:attn_vis}
\end{figure*}

Figure~\ref{fig:attn_vis} visualizes the cross-modal attention maps for affordance grounding, illustrating whether the model focuses on functionally relevant affordance regions prior to coordinate prediction or erroneously diffuses its focus over task-irrelevant background areas.

\begin{table}[ht]
\centering
\caption{OOD attention-alignment diagnostics on AGD20K.}
\label{tab:attn}
\resizebox{\linewidth}{!}{
\begin{tabular}{l c c c c c c c}
\toprule
Model & Size & gIoU$\uparrow$ & cIoU$\uparrow$ & P@50$\uparrow$ & KLD$\downarrow$ & SIM$\uparrow$ & NSS$\uparrow$  \\ 
\midrule
Qwen2.5-VL & 3B & 23.21 & 18.90 & 17.62 & 10.27 & 0.30 & 0.75  \\
RoboBrain2.0 & 3B & 27.52 & 22.38 & 23.67 & 13.31 & 0.18 & 0.58  \\

Qwen3-VL & 4B & 26.83 & 20.84 & 19.91 & \underline{9.29} & 0.34 & 0.91  \\
RoboBrain2.5 & 4B & 33.58 & 33.56 & 29.14 & 12.03 & 0.25 & 0.87  \\
RynnBrain & 4B & 27.11 & 25.21 & 21.04 & 10.16 & 0.32 & 0.83  \\
InternVL3 & 7B & 18.18 & 14.63 & 3.79 & 10.09 & 0.25 & 0.61 \\
Qwen2.5-VL & 7B & 25.70 & 23.82 & 18.22 & 10.13 & 0.32 & 0.85  \\
LISA & 7B & 13.18 & 11.96 & 1.45 & 13.68 & 0.16 & 0.46 \\
SAM4MLLM & 7B & 15.27 & 13.22 & 2.40 & 9.51 & 0.27 & 0.52 \\
Seg-Zero & 7B & 26.99 & 22.01 & 6.52 & \textbf{9.02} & 0.35 & 0.94 \\
Qwen3-VL & 8B & 26.74 & 29.44 & 19.44 & 10.37 & 0.31 & 0.88  \\
RoboBrain2.0 & 7B & 27.32 & 26.60 & 23.94 & 12.71 & 0.28 & 0.78  \\
Affordance-R1 & 7B & 31.38 & 27.85 & 20.49 & 9.73 & 0.36 & \underline{0.98} \\
RoboBrain2.5 & 8B & \textbf{36.82} & \textbf{49.07} & \underline{37.08} & 10.28 & \underline{0.38} & 0.89  \\
RynnBrain & 8B & 27.55 & 28.24 & 20.21 & 10.42 & 0.32 & 0.87  \\
\midrule
SpatialAfford & \textbf{4B} & \underline{36.21} &\underline{48.76} & \textbf{37.44} & 9.35 & \textbf{0.40}& \textbf{1.05}\\
\bottomrule
\end{tabular}
}
\end{table}
\subsection{Ablation: Component Synergy}
\label{sec:exp_ablation}

Table~\ref{tab:ablation} isolates the incremental contribution of each component. Standard SFT delivers limited gains on the baseline model (P@50 rises from 17.50\% to 22.50\%), whereas integrating Spatial Attention Alignment (SAA) substantially elevates P@50 to 40.50\% and gIoU to 42.59\%, confirming that explicit cross-modal attention regularization is critical for anchoring visual focus onto affordance regions. Standalone GRPO mainly optimizes coordinate prediction precision. Incorporating Spatial-Aware GRPO policy optimization achieves the best overall performance (47.80\% gIoU and 55.00\% P@50), demonstrating that reward-driven boundary refinement is most effective when executed over an aligned visual feature space. These progressive gains confirm the functional dependency between components: SAA establishes affordance-aware visual focus, which SpatialAfford subsequently transforms into precise spatial coordinates.

\begin{table}[ht]
\centering
\caption{Component ablation on ShareRobot.}
\label{tab:ablation}
\small
\resizebox{\columnwidth}{!}{%
\begin{tabular}{lccccc}
\toprule
Variant & SAA & GRPO & gIoU$\uparrow$ & cIoU$\uparrow$ & P@50$\uparrow$\\
\midrule
Baseline (Qwen3-VL-4B) & \xmark & \xmark & 25.10 & 16.84 & 17.50 \\
SFT & \xmark & \xmark & 27.17 & 45.59 & 22.50 \\
SFT + SAA & \cmark & \xmark & \underline{42.59} & \underline{47.16} & 40.50 \\
GRPO & \xmark & \cmark & 42.45 &  43.34  & \underline{45.55} \\
\midrule
\textbf{SpatialAfford} & \cmark & \cmark & \textbf{47.80} & \textbf{49.76} & \textbf{55.00}\\
\bottomrule
\end{tabular}%
}
\end{table}

We investigate the sensitivity of SAA to the loss weight $\alpha$ and the choice of loss function. Results show that KL divergence with $\alpha{=}0.1$ achieves the optimal trade-off, maintaining superior grounding ability (42.59\% gIoU, 40.50\% P@50) while effectively enhancing attention alignment.
In contrast, pure coverage loss severely degrades coordinate prediction despite improving attention alignment, justifying our adoption of KL divergence for SAA.

\begin{table}[ht]
\centering
\caption{$\alpha$ sensitivity and loss function selection on Qwen3-VL-4B.}
\label{tab:alpha_sweep}
\small
\resizebox{\columnwidth}{!}{%
\begin{tabular}{clccccccl}
\toprule
$\alpha$ & Loss Type & Loss Value & gIoU $\uparrow$ & P@50 $\uparrow$   \\
\midrule
0 & Baseline & --- & 25.10 & 17.68  \\
0.01 & kl divergence & 0.5455 & 35.26 & 33.25  \\
0.05 & kl divergence & 0.5005 & 37.49 & 36.68  \\
\textbf{0.1} & kl divergence & 0.5125 & \textbf{42.59} & \textbf{40.50} \\
0.2 & kl divergence & 0.5312 & 33.12 & 24.10  \\
0.1 & coverage & 0.4755 & 30.67 & 32.15 \\
1.0 & coverage & 0.3702 & 33.38 & 31.73 \\
\bottomrule
\end{tabular}%
}
\end{table}

%==============================================================================
\section{Conclusion}
%==============================================================================

This work shows that compact VLMs for affordance grounding should not be optimized only at the output level. By explicitly aligning visual attention before policy-based coordinate refinement, SpatialAfford builds a more reliable pathway from task intent to affordance regions. The empirical results suggest that improving where a compact model looks is a key prerequisite for robust affordance grounding, and that attention alignment and spatial policy optimization provide complementary benefits.Future work will extend SpatialAfford’s 2D attention alignment to 3D representations to establish view-consistent functional focus, bridging 2D semantic perception and 3D physical manipulation.

\clearpage

\bibliography{acm_grpo}

\end{document}